\documentclass[sigconf]{acmart}

\copyrightyear{2022}
\acmYear{2022}
\setcopyright{acmcopyright}
\acmConference[MMSports '22] {Proceedings of the 5th International ACM Workshop on Multimedia Content Analysis in Sports}{October 10, 2022}{Lisboa, Portugal.}
\acmBooktitle{Proceedings of the 5th International ACM Workshop on Multimedia Content Analysis in Sports (MMSports '22), October 10, 2022, Lisboa, Portugal}
\acmPrice{15.0}
\acmISBN{978-1-4503-9488-8/22/10}
\acmDOI{10.1145/3552437.3555705}

\AtBeginDocument{%
  \providecommand\BibTeX{{%
    \normalfont B\kern-0.5em{\scshape i\kern-0.25em b}\kern-0.8em\TeX}}}

\graphicspath{ {images/} }
\usepackage{xcolor}

\usepackage{subcaption}
\usepackage{longtable} 

\begin{document}

\title{Towards Automated Key-Point Detection in Images with Partial Pool View}


\author{Timothy Woinoski and Ivan V. Baji\'c}
\email{twoinosk@sfu.ca, ibajic@ensc.sfu.ca}
\orcid{1234-5678-9012}
\affiliation{
  \institution{Simon Fraser University}
  \streetaddress{8888 University Drive}
  \city{Burnaby}
  \state{BC}
  \country{Canada}
  \postcode{V5A1S6}
}

\renewcommand{\shortauthors}{Timothy and Ivan}

\begin{abstract}
Sports analytics has been an up-and-coming field of research among professional sporting organizations and academic institutions alike. With the insurgence and collection of athlete data, the primary goal of such analysis is to improve athletes' performance in a measurable and quantifiable manner. This work is aimed at alleviating some of the challenges encountered in the collection of adequate swimming data. Past works on this subject have shown that the detection and tracking of swimmers is feasible, but not without challenges. Among these challenges are pool localization and determining the relative positions of the swimmers relative to the pool.
This work presents two contributions towards solving these challenges. First, we present a pool model with invariant key-points relevant for swimming analytics. Second, we study the detectability of such key-points in images with partial pool view, which are challenging but also quite common in swimming race videos.
\end{abstract}



\maketitle


\section{Introduction} \label{sec:introduction}
Sports analytics has been an up-and-coming field of research among professional sporting organizations and academic institutions alike. With the insurgence and collection of athlete data, the primary goal of such analysis is to try and improve athletes' performance in a measurable and quantifiable manner. This goal is in contrast to traditional coaching methods in which a coach relies solely on experience and methods that seem to work well. A more ideal situation would be to have coaches utilize both experience and data to better direct the training of their athletes. This practice has started to appear in sports where adequate data is readily available \cite{booth2019mathematical, 3pointBoom}. Unfortunately, adequate data is hard to obtain and is generally not easily available in most sports. This work is concerned with the automated collection of swimming data in a competition setting.

Previous works on this topic~\cite{hall2021detection,woinoski2021swimmer} have shown that the detection and tracking of swimmers is possible, however, they each face their own challenges. For example,~\cite{hall2021detection} assumes that video of swimmers is captured by a static camera and that the entire pool is visible. This is generally not the case as the equipment and facilities to do so are expensive and limited. In~\cite{woinoski2021swimmer}, which does not assume a static camera, different drawbacks are observed. Long-term tracking of swimmers who leave the camera's field of view is difficult due to the challenges of re-identification. Furthermore, there is no trivial way to automatically map any collected swimmer analytics to any given swimmer in the field of view. To overcome such challenges, other automated analytics solutions~\cite{sportlogiq,hall2021detection}, introduce field localization as a method for producing results that are more robust to the mentioned issues. 

\begin{figure}[ht]
    \centering
    \begin{subfigure}[b]{0.495\linewidth}
        \centering
        \caption{Base pool model}
        \includegraphics[width=\linewidth,height=2.5cm]{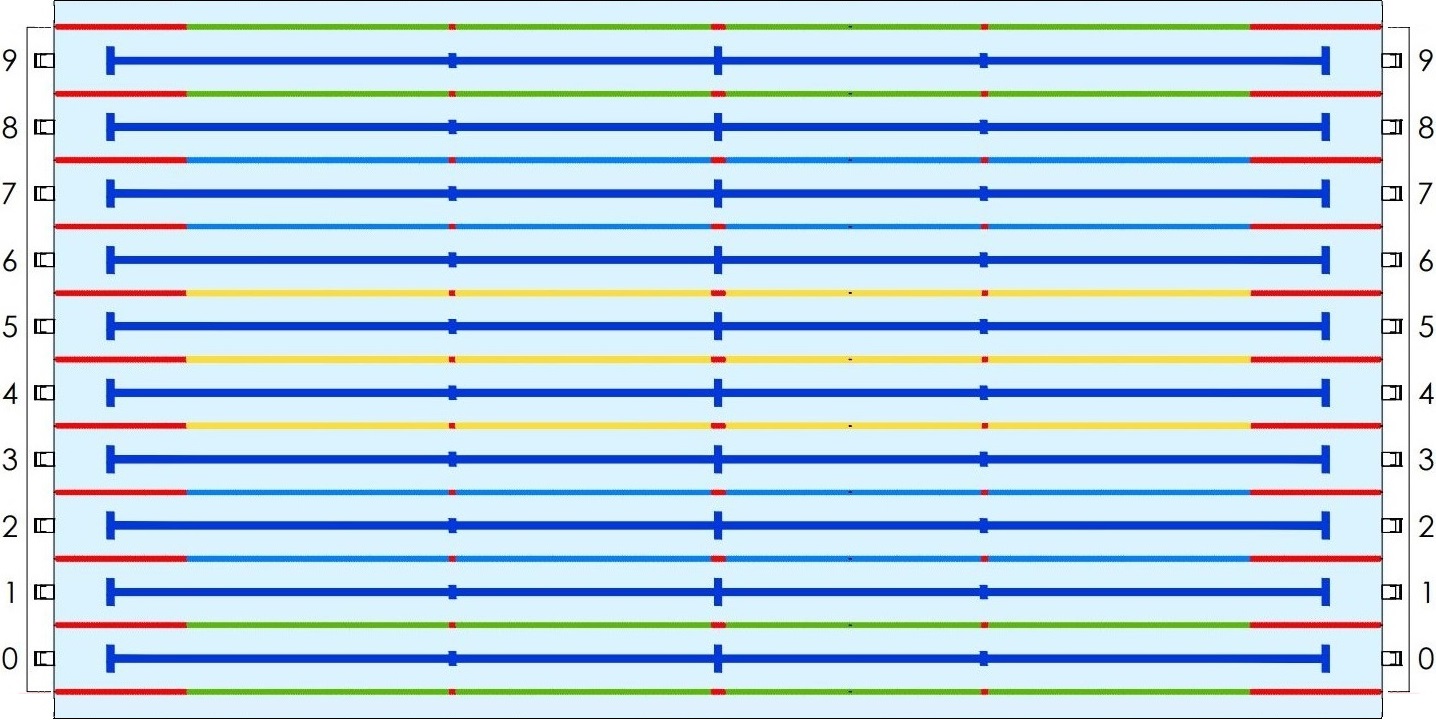}
        \label{fig:poolBaseHomography}
    \end{subfigure}
    \begin{subfigure}[b]{0.495\linewidth}
        \centering
        \caption{8$\times$50 Pool}
        \includegraphics[width=\linewidth,height=2.5cm]{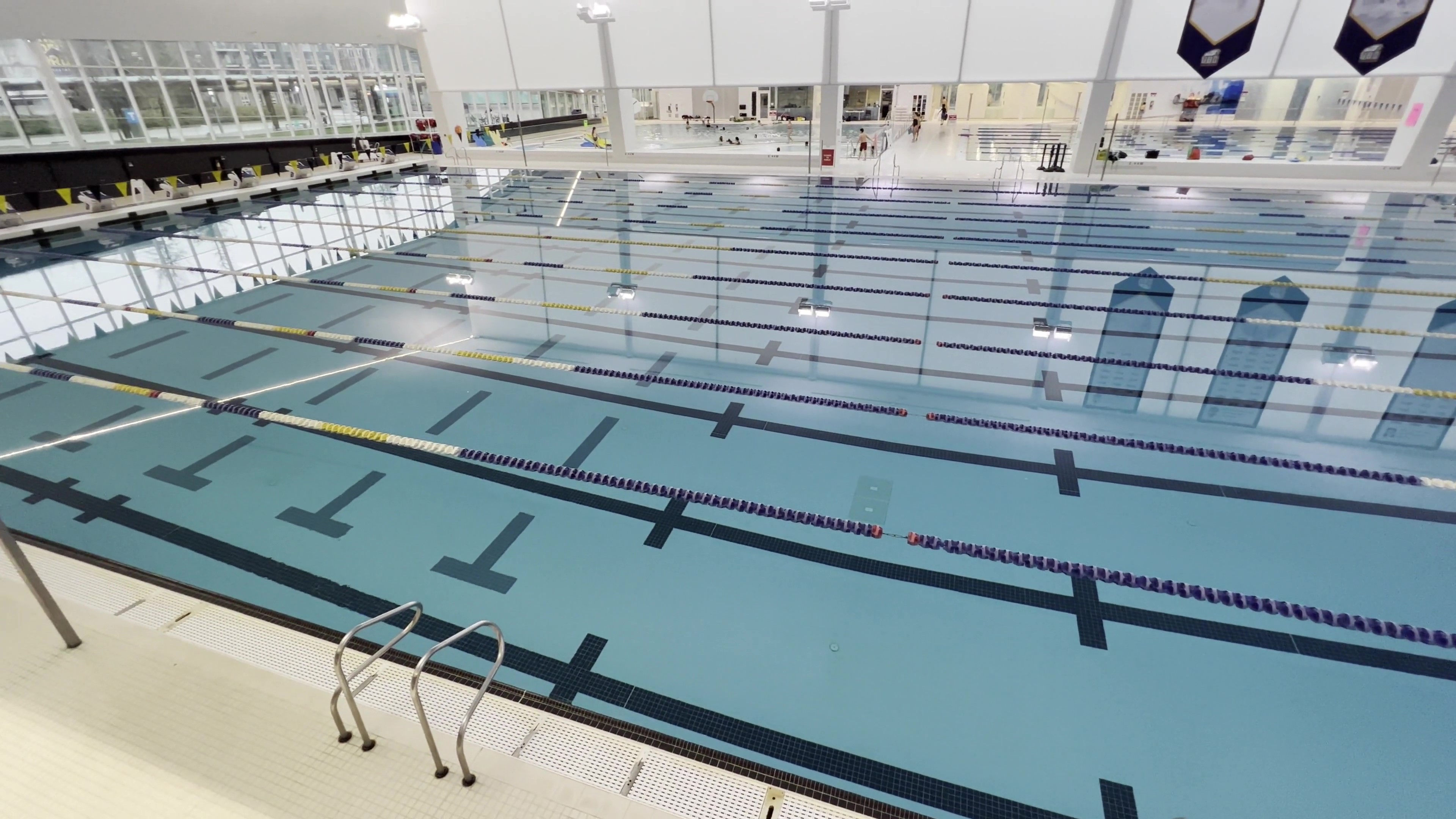}
        \label{fig:samplePoolIm}
    \end{subfigure}
    \begin{subfigure}[b]{\linewidth}
        \centering
        \caption{Sample image transformed by human generated homographic projection to fit over the base pool image.}
        \includegraphics[width=\linewidth]{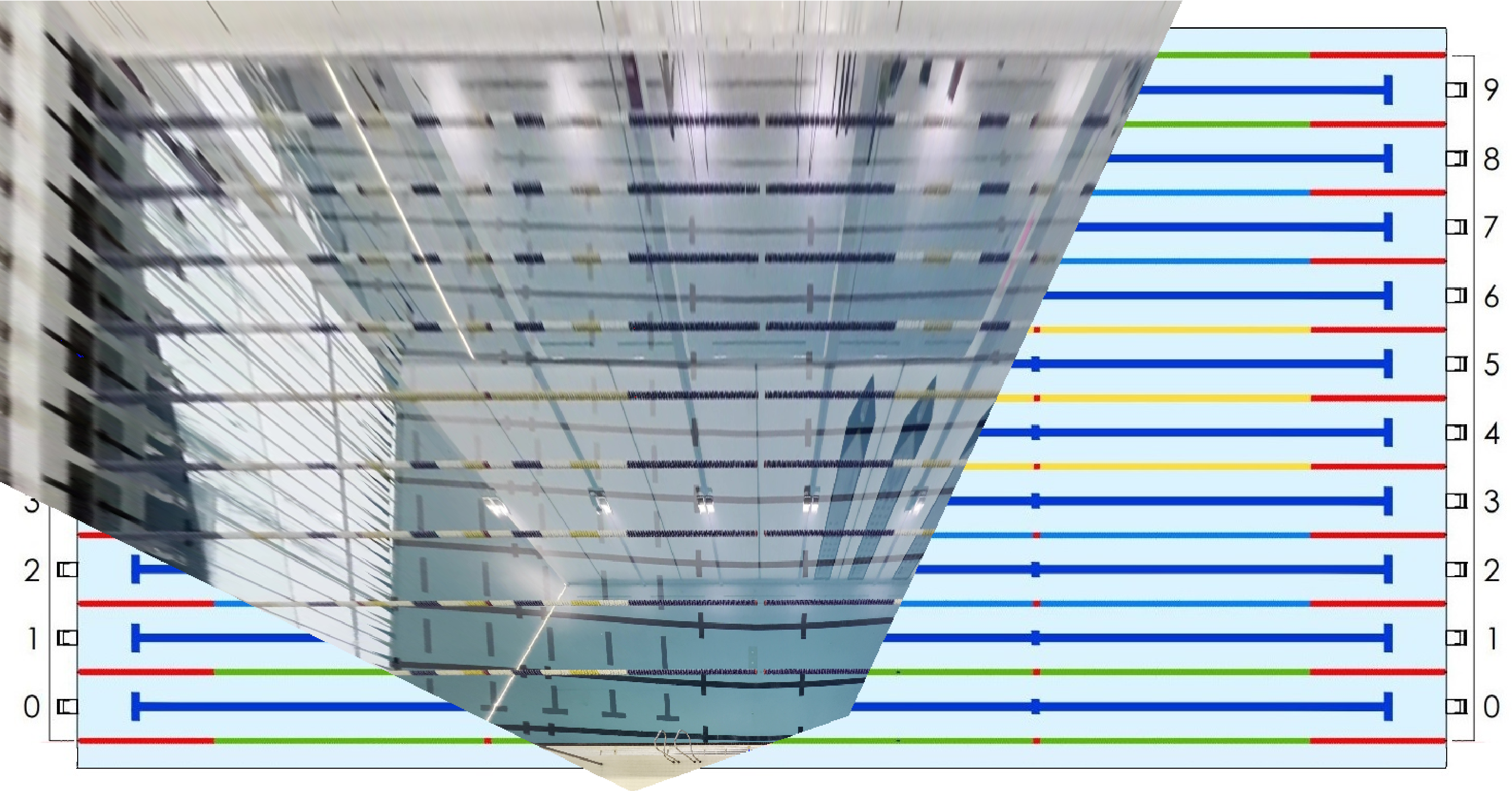}
        \label{fig:exampleHomography}
    \end{subfigure}
        \caption{An example of pool localization characterized by a simple homographic projection.}
        \label{fig:homographicProjection}
\end{figure}

In the context of this work, pool localization can be characterized by a homography that maps a given frame to a base frame. An example can be seen in Figure~\ref{fig:homographicProjection} where the given frame is seen in Figure~\ref{fig:samplePoolIm} and the base frame is seen in Figure~\ref{fig:poolBaseHomography}. The projection of the given frame onto the base image can be seen in Figure~\ref{fig:exampleHomography}.

Localization of the pool would allow a system to know what portion of the pool is being observed at any given time. If this is known, then the system would also know the position of any detected swimmer relative to the boundaries of the pool, and also the lanes in which they swim. With the successful completion of such a task, the mentioned problems can be overcome. This work presents two main contributions towards solving the above challenges, which we consider to be the beginnings of an automated pool localization method that can handle general pool images: 
\begin{enumerate}
    \item We present a pool model called \textit{base pool} with invariant key-points relevant for
swimming analytics.
    \item We study detectability of such key-points in images with partial pool view, by training a deep model for such key-point detection.
\end{enumerate}

This paper is structured as follows. First, an overview of related work is presented in Section~\ref{sec:relatedWork}. In Section~\ref{sec:methods}, the method of pool localization and the details related to reproducing the results are presented. Section~\ref{sec:results} goes over how well the methods worked, the meaning of the results given, and what can be done to improve the results. Lastly, some finial thoughts for moving forward are presented in Section~\ref{sec:conclusion}.


\section{Related Work} \label{sec:relatedWork}
There are many methods for localization described in the literature, but most of those are geared toward object detection and localization~\cite{cadena2016past}. The problem of sports field localization is more specific, in the sense that many properties of the field to be localized are known \textit{a priori}; as a consequence, more assumptions can be made allowing for more elaborate solutions. Broadly, previous work on sports field localization can be divided into 
the following two categories: hand-crafted methods, and deep-learning-based solutions. 

There are many well-defined methods for extracting the lines, circles and ellipsoids that make up sports fields utilizing traditional image processing methodology~\cite{GonzalezWoods2018}. As a result, there are many hand-crafted methods for localizing a sports field 
that build upon such methods~\cite{cuevas2020automatic,sharma2018automated,hadian2015fast,brown2007automatic}.The approach in~\cite{brown2007automatic} 
utilizes SIFT features~\cite{lowe2004distinctive} in combination with the RANSAC algorithm~\cite{fischler1981random} as a baseline method to compare to. The methods in~\cite{cuevas2020automatic,hadian2015fast} utilize a myriad of different classical processing methods to extract the points and lines of the field of play and then use these extracted lines and points to produce a homography~\cite{hartley2004}, which effectively localizes the field of play. It should be noted that~\cite{sharma2018automated} utilizes deep learning to match a segmentation map in a dictionary of maps that define a particular homography, by which 
the image being considered is localized. Thus, this method could also be categorized as a deep learning-based solution. However, 
the traditional image processing methods also proposed in their work were more successful than the deep learning-based counterpart. 
Many of the mentioned works produced very respectable results and thus give a strong argument for approaching the problem of pool localization with hand-crafted methodology.

Deep learning has become very popular in the last decade. Accordingly, there are many options for applying deep learning models to solve large-scale problems in computer vision. 
Works from~\cite{homayounfar2017sports,fani2021localization,nie2021robust,citraro2020real} apply a variety of 
methodologies that rely on deep learning to do the brunt of the localization work. The approach in~\cite{homayounfar2017sports} is considered by some as one of the first machine learning-based methods for sports field localization utilizing deep learning. They implement a segmentation network that separates the field pixels from the non-field pixels. Once completed, the resulting segmentation is utilized by another loss-function-based system to predict the vanishing points of the two sets of parallel lines that make the field boundaries of the field in the image in question. Once the vanishing points are calculated, the field can be characterized by a homography, and thus is localized. The work presented in~\cite{fani2021localization} is a comprehensive report on how to robustly localize a sports field from broadcast video, which contains many different views (zoomed in and out), and commercial breaks. The 
data produced for this broadcast video was homography transform parameters for each frame in the video to a base field model. The model employed in their solution was trained to take a frame and produce a vector that characterizes the frame's homography. Lastly, the work reported in~\cite{nie2021robust,citraro2020real} relies on the detection of so-called key-points that are generally unique and represent a point whose location is known in both the base model 
and the image in question. With enough key-points correctly detected in a given image, a homography can be computed, and thus, the given image is localized. Both works rely on variations of the widely utilized U-Net~\cite{ronneberger2015u} to produce segmentation maps or volumes in which each channel is associated with one key-point and represents the probability of finding that key-point in the input image. The point with the highest probability is chosen as the predicted location of the corresponding key-point. Once again, the above mentioned works perform very well and make the selection of a methodology for pool localization difficult. 

Pool localization is a special case of general sports field localization. 
However, to our knowledge, there is no work currently available  that automatically localizes a swimming pool given an image with partial pool view, i.e., an image where only a portion of the pool is observed. The only known (to us) reference on the topic is~\cite{hall2021detection}, however, it focuses on pool localization in images of the entire pool, from a calibrated static camera with known internal and external camera parameters. 
Besides this, the topic of swimming pool localization 
is relatively untouched. Unfortunately, this also means that there is a severe lack of data to be utilized for research.


\section{Methods} \label{sec:methods}
Given the large body of related work on other sports, a key-point detection methodology similar to~\cite{nie2021robust,citraro2020real} is chosen as the preferred method of localization in this work. To implement such a method, the appropriate data is required. As key-point detection is being implemented, a model must be proposed such that consistent key-points are collected for any given pool. In addition, images of pools must be obtained that are sufficiently different such that the model can learn to generalize the detection of proposed key-points. Once data is available, a deep learning model can be constructed and trained. With such a trained  model, the detectibility of the proposed key-points can be approximated by considering the key-point detection performance of the created model.

\begin{figure*}[ht]
    \centering
    \includegraphics[width=\linewidth]{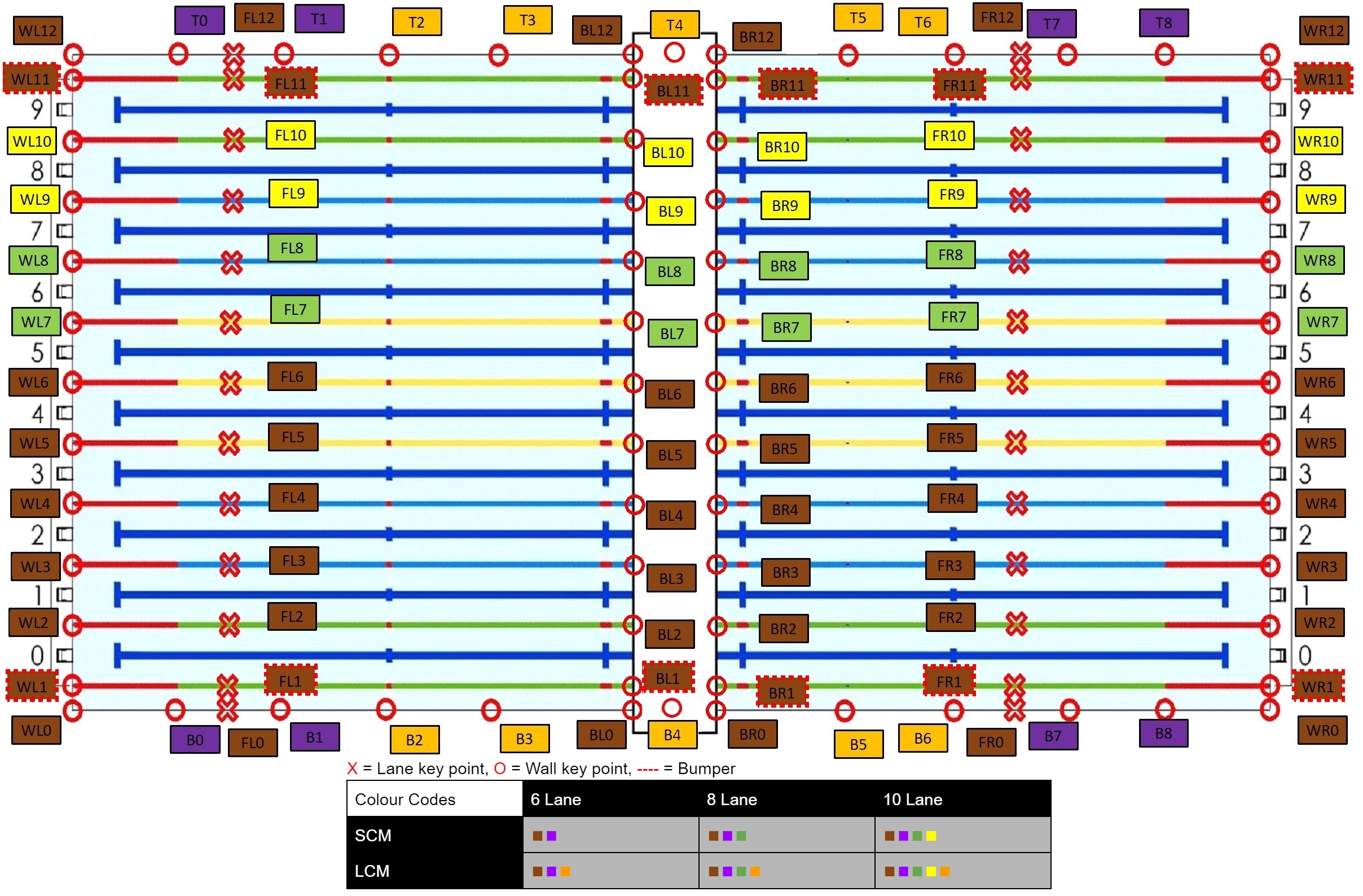}
    \caption{Proposed pool model and respective key-point locations.}
    \label{fig:poolModel}
\end{figure*}


\begin{table*}[ht]
    \centering
    \begin{tabular}{|p{0.12\linewidth}|p{0.05\linewidth}|p{0.83\linewidth}|}
        \hline
        \textbf{key-points} & \textbf{KP\#} & \textbf{Description}\\
        \hline
        Wall Left & [1, 11] & Defined as the intersection of the number lane-rope and the left wall. 1 and 11 don't exist at times.\\
        \hline
        Wall Left & 0 \& 12 & Defined as the bottom left corner and top left corner of the pool respectively.\\
        \hline
        Wall Right & [1, 11] & Same as wall left but on the right side of the pool.\\
        \hline 
        Wall Right & 0 \& 12 & Defined as the bottom right corner and top right corner of the pool respectively.\\
        \hline
        Floating Left & [1, 11] & Defined as the intersection of the left side of a number lane-rope and the edge of the frame. 1 and 11 don't exist at times.\\
        \hline 
        Floating Left & 0 \& 12 & Defined as the intersection of the left side of the bottom and top walls, respectively, and the edge of the frame.\\
        \hline
        Floating Right & [1, 11] & Same as floating left but on the right side of the pool.\\
        \hline 
        Floating Right & 0 \& 12 & Same as floating left but on the right side of the pool.\\
        \hline
        Bulkhead Left & [1, 11] & Defined as the intersection of the right side of a number lane-rope and an existing bulkhead. 1 and 11 don't exist at times.\\
        \hline
        Bulkhead Left & 0 \& 12 & Defined as the intersection of the right side of the bottom and top walls respectively, and an existing bulkhead.\\
        \hline
        Bulkhead Right & [1, 11] & Defined as the intersection of the left side of a number lane-rope and an existing bulkhead. 1 and 11 don't exist at times.\\
        \hline
        Bulkhead Right & 0 \& 12 & Defined as the intersection of the left side of the bottom and top walls respectively, and an existing bulkhead.\\
        \hline
        Wall Top & [0, 8] & Defined every 5m of the length of the pool on the top wall. T4 is not present when bulkheads are present.\\
        \hline
        Wall Bottom & [0, 8] & Same as wall top but on the bottom wall of the pool.\\
        \hline
    \end{tabular}
    \caption{Summarizes the location of key-points in Figure~\ref{fig:poolModel}}
    \label{tab:keyPointDefinitions}
\end{table*}

\subsection{Base Pool Model}\label{subsec:poolModel}
The base pool model, seen in Figure~\ref{fig:poolModel}, is what defines where and what key-points should be identified in any given image. Humans are very good at recognizing objects in their environment that are the same across different points of view, unfortunately, computers are not. A set of possible points, must be defined such that they are consistently recognized and learned by a key-point detecting algorithm. 

When constructing this key-point set, one must consider what constitutes a good key-point and what types of pools should such key-points be defined. In this work, we consider 9 different types of pools which can be categorized by two numbers in the following notation ``$n \times m$'', where $n$ is the number of lanes and $m$ is the length of the pool. $m$ can take values of 25 and 50, known as short course meters (SCM) and long course meters (LCM) respectively. $n$ represents the number of lanes in the pool being localized and can take values of 6, 8, 10, 12, 16, and 20, $n$ can only be greater than 10 if $m$ is SCM. When pools have $n$ values of 12, 16, and 20 they contain a bulkhead, seen in Figure~\ref{fig:UBCRightRes}, and~\ref{fig:UBCLeftRes} which separates one sub-pool from another; this is a common occurrence in SCM competitions. Given the possible pool types, we propose the following pool model seen in Figure~\ref{fig:poolModel} for which Table~\ref{tab:keyPointDefinitions} gives definitions of each key-point location. 

The image in Figure~\ref{fig:poolModel} defines 96 different key-points which are unique within any pool setting considered in this work. For ease of communication, each of these 96 key-points can be referred by one into one of the following classes, wall left, right, top, and bottom, bulkhead left and right, and finally floating right and left. All key-point classes have numbers, either in the range of $[0, 12]$ for classes that represent lanes or $[0, 8]$ for classes that represent the length of the pool. Pools also tend to differ in the number of lane-ropes\footnotemark{} that divide the pool, for example an ten lane pool can have 9, 10, or 11 lane-ropes. As such the key-points marked as bumpers seen in figure~\ref{fig:poolModel} are explicitly considered as points that may more may not exist. Bumpers are defined as the lane-ropes that divided the wall from the outside lanes. In Figure~\ref{fig:SaanichRes} there are bumpers separating lanes 8 and 1 from the wall top and bottom. The floating key-points are different than traditional key-points utilized for homography creation. When considering their location in the base frame they are invariant in the vertical direction only, that is, there is infinitely many locations in the horizontal direction that a floating point may be. If you look carefully, each key-point that has the same class is roughly mutable with another. That is, for example, wall left $2$ is identical to wall left $3$. What differentiates mutable key-points is their location relative to each other key-point in that class and their count relative to the top and bottom walls. Note the locations and numbers of specific key-points are chosen such that they are most similar across all different pools. This is to allow the detection model to have an easier time recognizing similar key-points in different pools. These thoughts considered, there is likely an innumerable number of ways to select key-points and their locations, the proposed model is only one of those such enumerations. Lastly, while it is not known if the chosen key-points are optimal in terms of detectability, they are essential in that they allow for robust characterization of the pool in a given image.

\footnotetext{A lane-rope is a rope running the length of a pool, separating each lane from one another or the wall, if a lane-rope is separating the wall from a lane it is known as a bumper.}

\subsection{Data}
With the pool model defined, the next step is to collect images of pools for localization. In an ideal situation, examples suitable for training models are independent and identically distributed. However, as with most deep learning training data, collecting example images is not trivial, competition pools are not plentiful in a given area, they are generally spread over large distances, and as such, this limited the number of pools that could be utilized for this work. With these details noted, video footage from five pools, in various configurations, was collected. Unfortunately, each frame from a video is highly dependent on the next, to mitigate this issue, images taken from each video are sampled every 15-30 frames, depending on how the video was collected. In addition, all video footage was taken from a minimum of three maximally different viewpoints in the pool, which increased the independence of all collected images from one pool. The collected video is landscape and portrait with a minimum resolution of 1080p (16x9) and at 30 frames per second. Lastly, all the videos showcase all nine pool types mentioned in Section~\ref{subsec:poolModel}. The images were annotated utilizing the Computer Vision Annotation Tool (CVAT)~\cite{CVAT}. The amount of data collected and the type can be viewed in Table~\ref{tab:dataDistabution}. Three main pool categories are depicted in this table, pools with six, eight, and ten lanes. Please note that an eight lane pool can have eight or 16 lanes depending on if there is a bulkhead present in the images or not; the same goes for six and ten lane pools. In total, 1,352 frames were used for training and 284 frames were used for testing.

\begin{table}[ht]
    \centering
    \begin{tabular}{|c|c|c|}
        \hline
        \textbf{Pool Type} & \textbf{Number Images Taken} & \textbf{Data-set}\\
        \hline
        20x25 & 48 & Test\\
        \hline
        20x25 and 10x50 & 234 & Training\\
        \hline
        16x25 and 8x50 & 180 & Test\\
        \hline
        16x25 and 8x50 & 899 & Training\\
        \hline
        6x25 & 56 & Test\\
        \hline
        6x25 & 119 & Training\\
        \hline
    \end{tabular}
    \caption{Summary of data used for training and testing}
    \label{tab:dataDistabution}
\end{table}

\subsection{Key-point Detector}
The key-point detector utilized in this work is a slight variation of the popular U-net~\cite{ronneberger2015u}, utilized by~\cite{citraro2020real} and~\cite{nie2021robust} which incorporates a res-net style encoder to the u-net architecture. Like in the mentioned works, the output of the key-point detector will be a volume $V\in\mathbb{R}^{M \times N \times C}$ such that $M$ and $N$ is the resolution of the input and $C$ is the total number of possible key-points, that is 96 in this pool model. It is important to note that the detector has no idea that each key-point has a class, the classes are defined simply for eases of communication. The model will be trained to predict a distribution for each channel $C$ in $V$ such that each channel encodes the position of a predefined key-point. This distribution is forced by making the prediction layer of the model the soft-max activation function~\cite{Goodfellow2016}. If a key-point is not present in a given input the associated target distribution for the corresponding channel $C$ is flat. If a key-point is present in a given input the associated target distribution is a delta function at the location of the key-point in the frame. In contrast to~\cite{ronneberger2015u} the output volume will be the same size as the input, this is because many key-points in the input frames are found at the edges of the input. This modification to u-net is implemented by zero-padding all convolutions such that they do not have a reduced output resolution. The described model is implemented utilizing the Tensorflow Keras~\cite{tensorflow2015} package.

\subsubsection{Detection Accuracy and Optimization Loss}
To train and quantifiably evaluate the key-point detection model, the following equations must be defined. Because the key-point detection model was predicting distributions, the optimization function utilized in this work is the cross-entropy loss~\cite{Goodfellow2016}, defined in Equation~\ref{equ:crossEntropyLoss}, where $y_i\in\mathbb{R}^{M \times N}$ is the target and $h_i \in\mathbb{R}^{M \times N}$ is the predicted channel. In this implementation, the loss of each channel in the volume is summed with equal weight to create the final loss function for a given input image.

\begin{equation} \label{equ:crossEntropyLoss}
    L(y, h) = -\sum_{j=1}^{C}\sum_{i=1}^{N \times M} y_i^{(j)} \ln h_i^{(j)}
\end{equation}

The accuracy of detecting key-points for a given input is defined by Equation~\ref{equ:f1}, which is the harmonic mean of the precision and recall (Equations~\ref{equ:precition}, and~\ref{equ:recall}) of the detections produced for a given image~\cite{Goodfellow2016}. In Equations~\ref{equ:precition} and~\ref{equ:recall} $tp$ refers to the number of true positives, that is key-points detected correctly by the model. $fp$ is the number of false positives or the number of times the model thought there was a key-point when there actually was not or the model predicted one but its location was predicted incorrectly. Lastly, $fn$ is the number of times the model did not predict a key-point when in fact it should have. For all mentioned equations, the output is in the range of $[0, 1]$ and the closer to $1$ the result is, the better the performance. To obtain the accuracy across a set of images the $F1$ score of each image is averaged.

\begin{equation} \label{equ:f1}
    F1 = 2 \cdot \frac{recall \cdot precision}{recall + precision}
\end{equation}

\begin{equation}  \label{equ:precition}
    precision = \frac{tp}{tp+fp}
\end{equation}

\begin{equation}  \label{equ:recall}
    recall = \frac{tp}{tp+fn}
\end{equation}

\subsubsection{key-point Detection Training}
key-point detection training was implemented utilizing the Tensorflow Keras API~\cite{tensorflow2015}. The training procedure was a standard training pipeline utilizing a training and test data set. The training deviated from the standard procedure in terms of memory management, which had to be considered to deal with the large tensors associated with the key-point detection methodology. This is because the input images had a resolution of $1080 \times 1920$, therefore, propagating tensors through the network and even creating the expected target volume $V\in\mathbb{R}^{M \times N \times C}$ of floating-point numbers requires a lot of memory. To deal with this problem the input images were scaled down by a factor of $3.75$. 
In addition to the reduced resolution, the batch size of each epoch was set to one. This is also due to memory issues, but also because the images are different in resolutions. 
Because of the small data set, some image augmentation was implemented in the form of random contrast augmentation. Other augmentation methods were not attempted as the position of the labels is related to their key-point definition. Accordingly, simply applying augmentation methods that changed the position of a key-point would not make sense by definition of some key-points. Lastly, the optimizer utilized in the training process was the Adam Optimizer~\cite{tensorflow2015}, which was given a learning rate of $1e$-$4$.

\subsubsection{Detecting Predicted key-points} \label{subsec:detectingPredictedKeyPoints}
Unlike the work in~\cite{nie2021robust} and~\cite{citraro2020real}, which utilized a confidence channel as one of the channels in the volume $V$ to determine which channels have key-points. This work measures the entropy, which is defined in Equation~\ref{equ:entropy}, of each output channel. Where $y\in\mathbb{R}^{M \times N}$ is a distribution of a particular channel $C$ of resolution $M \times N$, being equal to the input resolution. In each distribution the model gives the highest values to the location where it thinks the key-point corresponding to that channel is located in the frame.

\begin{equation} \label{equ:entropy}
    H(y) = -\sum_{i=1}^{M \times N} y_i \ln y_i 
\end{equation}





Because each channel represents a distribution, if the entropy of a channel is lower than a flat distribution multiplied by a constant $\beta \in [0,1]$, that is $H(y) < \beta \ln(N \cdot M)$, then it can be considered to be predicting the key-point it represents. 
The location of the key-point it represents is the location of the maximum value in the output tensor. 


\section{Results} \label{sec:results}
This section gives a summary of how the key-point detector performed. Firstly, the training is discussed, then the detection accuracy, and then a discussion of the reported results is presented.

\subsection{Training} \label{subsec:trainingResults}
\begin{figure}[ht]
    \centering
    \includegraphics[width=\linewidth]{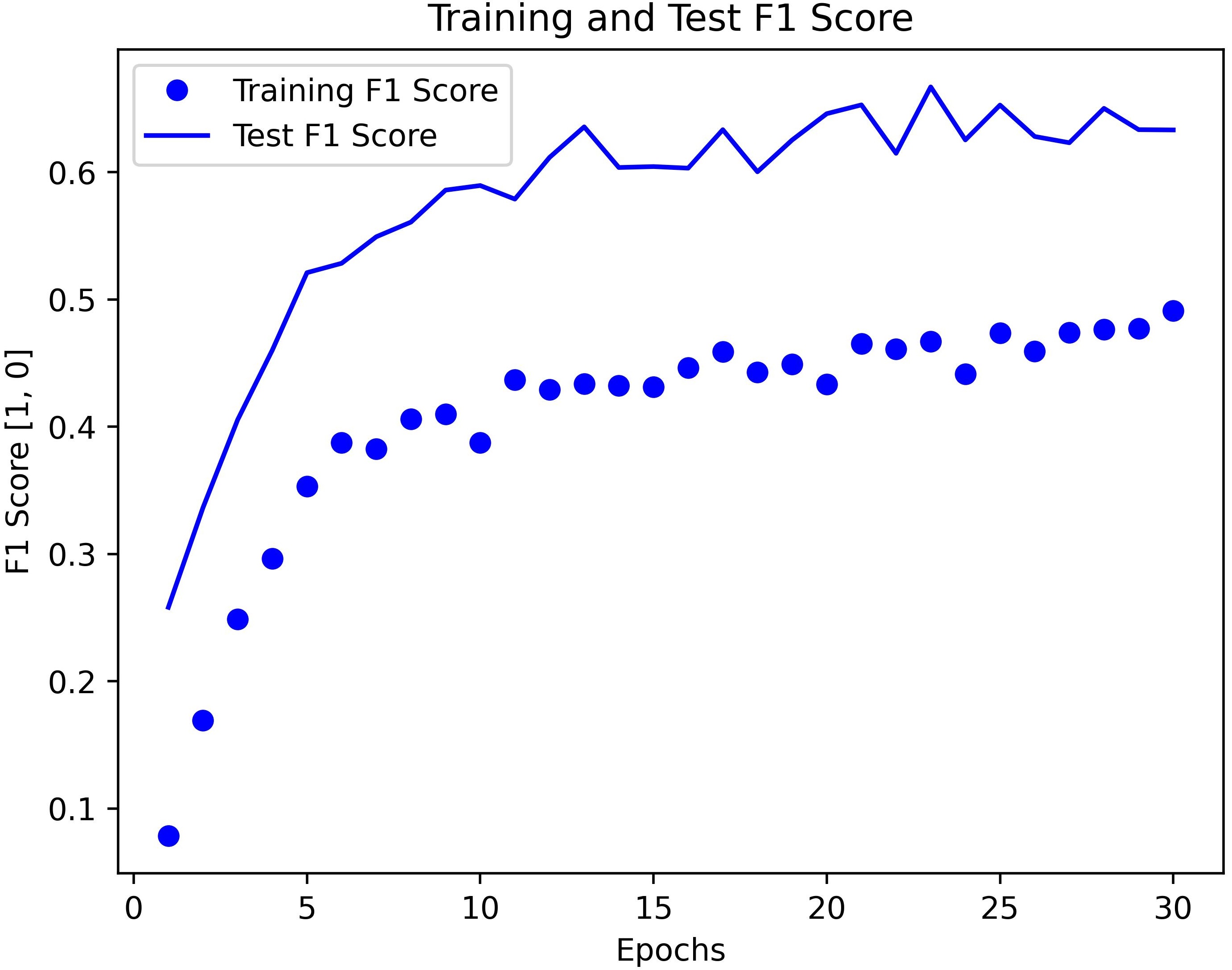}
    \caption{Training Plot}
    \label{fig:accuracyPlot}
\end{figure}

Reported in Figure~\ref{fig:accuracyPlot} is the per-frame average accuracy over the entirety of each test sequence as a function of epochs. It is worth noting that many different numbers of epochs were tried however it seemed that after 30 epochs the accuracy levels off.


\subsection{Accuracy Results} \label{subsec:accuracyResults}
\begin{figure}[ht]
    \centering
    \includegraphics[width=\linewidth]{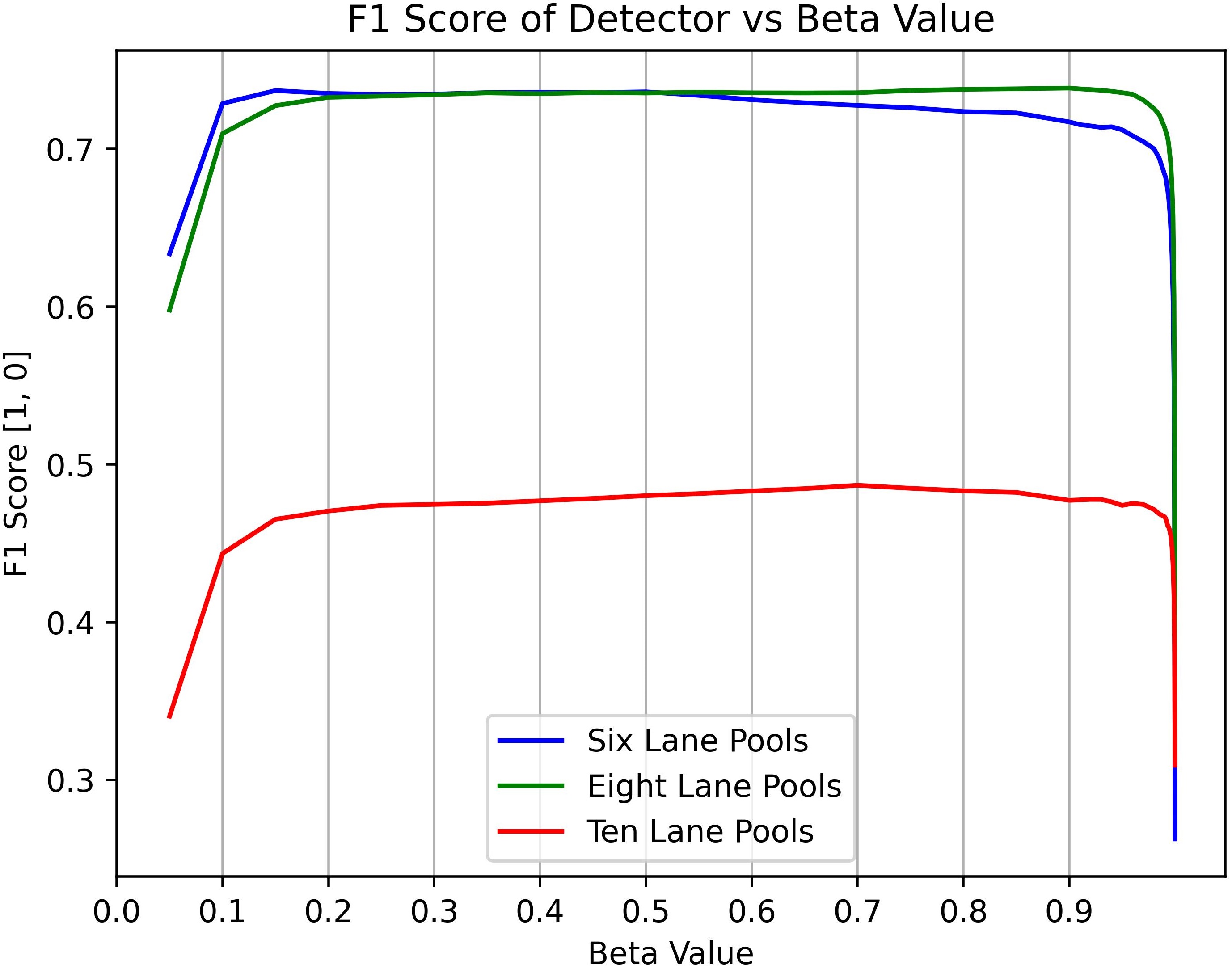}
    \caption{Average accuracy (F1) of the test sets vs different values of $\beta$, the control parameter selecting how confident the detector must be for a key-point prediction to be considered predicted. A prediction is considered correct if it is with five pixels of the ground truth.}
    \label{fig:betaAccuracyPlot}
\end{figure}

\begin{figure}[ht]
    \centering
    \includegraphics[width=\linewidth]{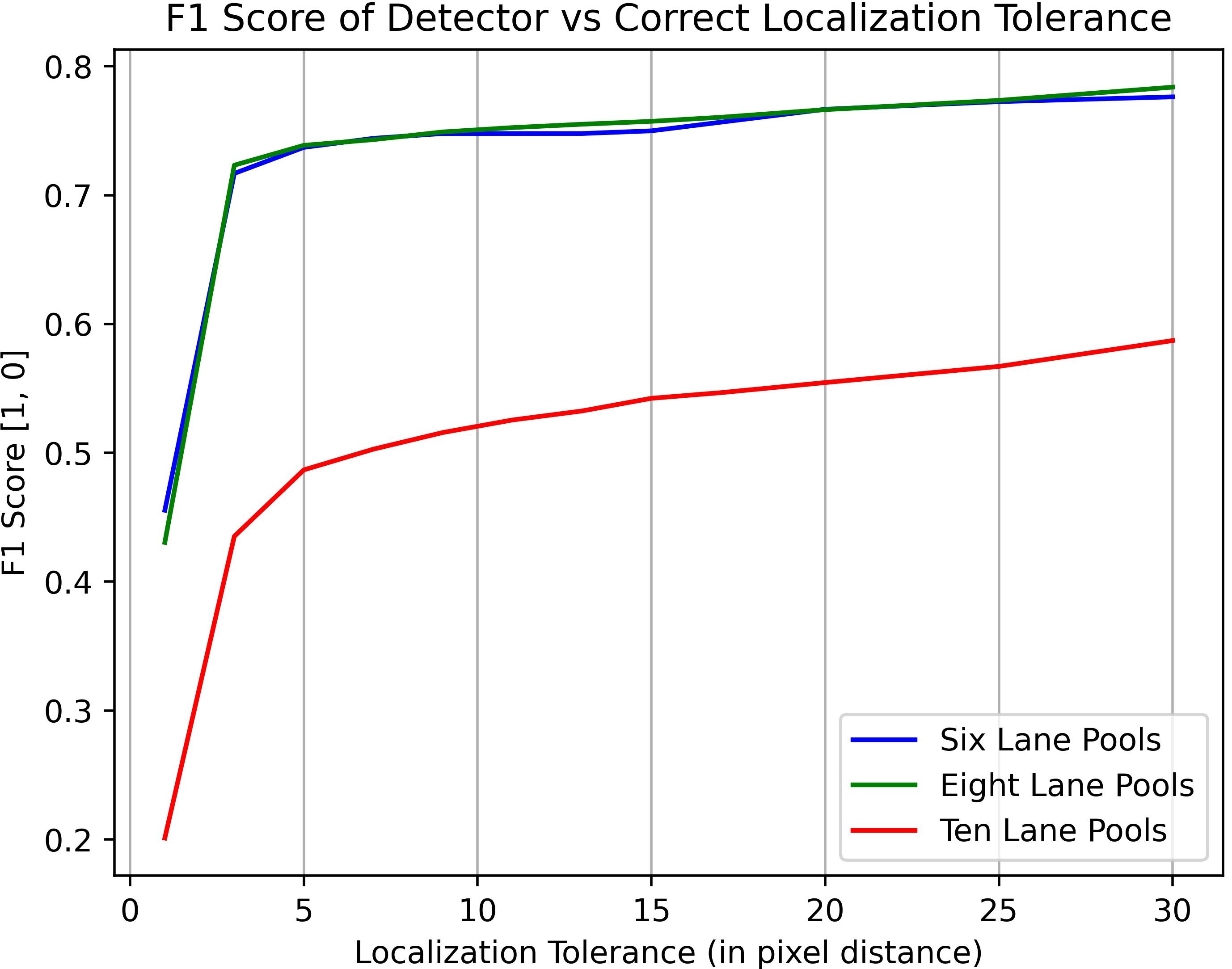}
    \caption{Average accuracy (F1) of the test sets vs different correct localization tolerance values. The $\beta$ value is set at the optimal value of beta for each type of pool, that is, $0.15$, $0.9$, and $0.7$, for six, eight, and ten lanes respectively. }
    \label{fig:pixTolerance}
\end{figure}

In this section, the accuracy of key-point detection is presented. Figure~\ref{fig:betaAccuracyPlot} details the per-frame average F1-Score over pools with different numbers of lanes as a function of $\beta$ for which the correct pixel tolerance is five pixels. Seen in Figure~\ref{fig:pixTolerance}, the per-frame average F1-Score over pools with different numbers of lanes as a function of pixel tolerance. Table~\ref{tab:keyPointResults} shows the precision, recall, and F1-Score of the test sequences, for each key-point class the model can predict. The ``Total'' column of the table reports the number of total key-points in the corresponding class that could be detected by the model. This column is necessary as some key-points are simply not present in some testing sequences. When this is the case the corresponding row has a value of zero and ``-''. In both Figure~\ref{fig:pixTolerance} and Table~\ref{tab:keyPointResults}, each pool type is given an optimal beta value based on the results seen in Figure~\ref{fig:betaAccuracyPlot}. Furthermore a pixel tolerance of five was chosen in this table for two reasons. Firstly, the chosen key-point detection method is very similar to the one presented in~\cite{citraro2020real}, in which a pixel tolerance of five was also chosen for input images of the same size or smaller. Secondly, anything less than five pixels would start to introduce noise in the key-point data. This is because key-points that are physically close to the camera take up substantially more pixels than key-points farther away. As such it is unclear what constitutes the exact position of a key-point. Lastly, Figure~\ref{fig:visualResults} gives five example images from the testing sequences to give a visual of how the model performs on the input data.

In Appendix~\ref{sec:AllData}, Table~\ref{tab:allData} gives the performance of the detector similar to Table~\ref{tab:keyPointResults} however, each key-point is broken down such that the performance of each point can be observed.


\begin{table}[ht]
    \centering
    \begin{tabular}{|l|c|c|c|c|}
        \hline
        \textbf{Class} & \textbf{Precision} & \textbf{Recall} & \textbf{F1}& \textbf{Total}\\
        \hline
        \multicolumn{5} {|c|} {\textbf{6 Lanes $\beta=0.15$}}\\
        \hline
        Wall Left & 0.1556 & 0.0726 & 0.0990 & 97\\
        \hline
        Wall Right & 0.2778 & 0.0494 & 0.0839 & 107\\
        \hline
        Floating Left & 0.9192 & 0.8621 & 0.8897 & 407\\
        \hline
        Floating Right & 0.9712 & 0.9118 & 0.9406 & 397\\
        \hline
        Bulkhead Left & - & - & - & 0\\
        \hline
        Bulkhead Right & - & - & - & 0\\
        \hline
        Wall Top & 0 & 0 & 0 & 74\\
        \hline
        Wall Bottom & 0 & 0 & 0 & 64\\
        \hline
        \multicolumn{5} {|c|} {\textbf{8 Lanes $\beta=0.9$}}\\
        \hline
        Wall Left & 0.5683 & 0.8301 & 0.6747 & 311\\
        \hline
        Wall Right & 0.7105 & 0.7892 & 0.7478 & 346\\
        \hline
        Floating Left & 0.7756 & 0.8941 & 0.8307 & 1510\\
        \hline
        Floating Right & 0.7901 & 0.9112 & 0.8463 & 1459\\
        \hline
        Bulkhead Left & 0.0580 & 0.2801 & 0.0961 & 377\\
        \hline
        Bulkhead Right & 0.0220 & 0.0909 & 0.0355 & 386\\
        \hline
        Wall Top & 0.3009 & 0.7893 & 0.4357 & 350\\
        \hline
        Wall Bottom & 0.1277 & 0.2530 & 0.1697 & 251\\
        \hline
        \multicolumn{5} {|c|} {\textbf{10 Lanes $\beta=0.7$}}\\
        \hline
        Wall Left & 0.0929 & 0.2682 & 0.1380 & 183\\
        \hline
        Wall Right & 0.4444 & 0.5140 & 0.4767 & 217\\
        \hline
        Floating Left & 0.8235 & 0.8435 & 0.8333 & 315\\
        \hline
        Floating Right & 0.7350 & 0.8741 & 0.7986 & 297\\
        \hline
        Bulkhead Left & 0.1268 & 0.3708 & 0.1890 & 338\\
        \hline
        Bulkhead Right & 0.2161 & 0.3942 & 0.2791 & 358\\
        \hline
        Wall Top & 0.1996 & 0.2450 & 0.2200 & 301\\
        \hline
        Wall Bottom & 0 & 0 & 0 & 86\\
        \hline        
    \end{tabular}
    \caption{Key-point class accuracy for pools with different numbers of lanes, with a pixel tolerance of five pixels.}
    \label{tab:keyPointResults}
\end{table}

\begin{figure}[ht]
    \centering
    \begin{subfigure}[b]{.49\linewidth}
        \centering
        \caption{16x25 F1 = 0.766}
        \includegraphics[width=\linewidth]{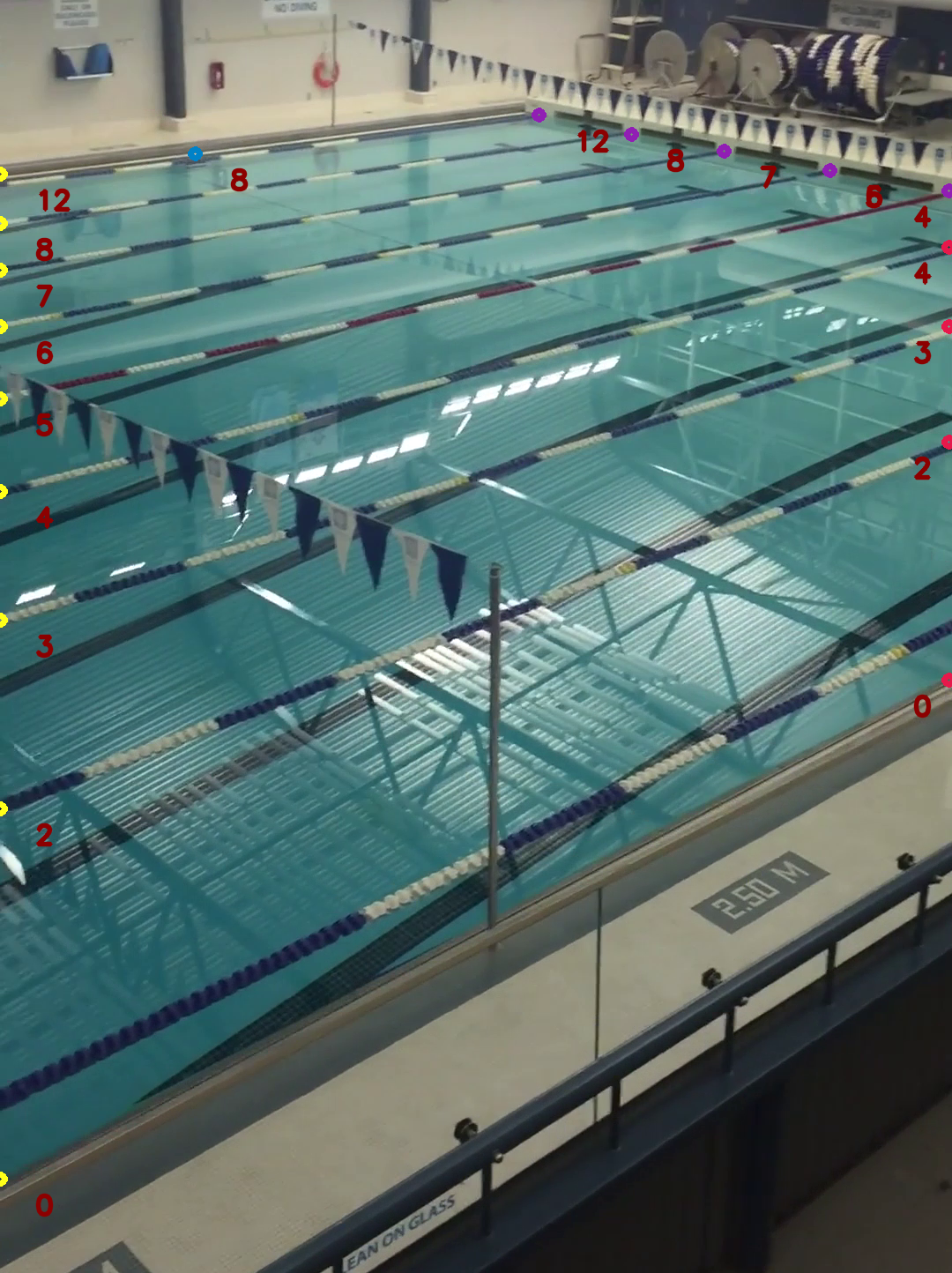}
        \label{fig:CGACLondonRes}
    \end{subfigure}
    \begin{subfigure}[b]{.49\linewidth}
        \centering
        \caption{6x25 F1 = 0.750}
        \includegraphics[width=\linewidth]{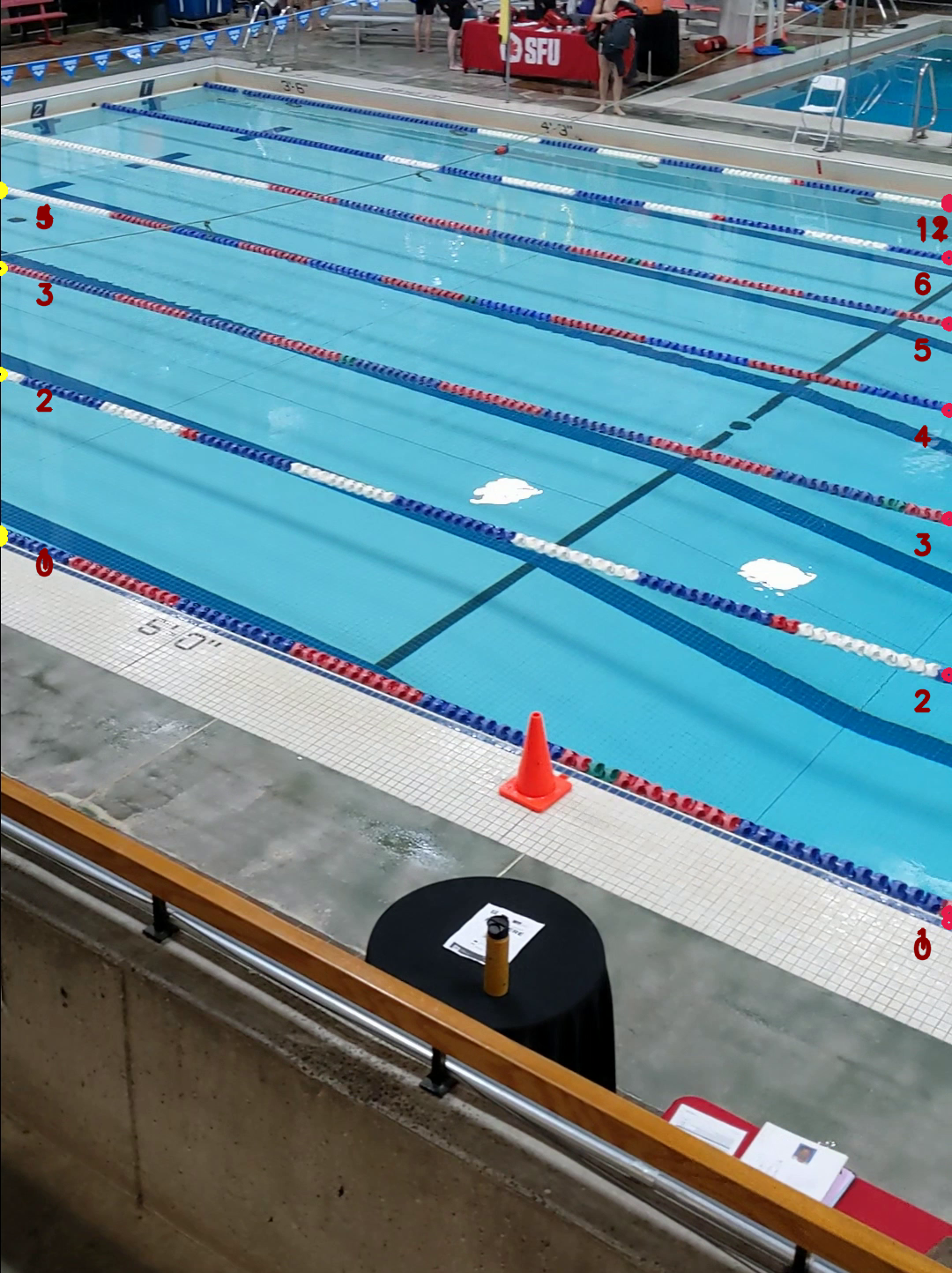}
        \label{fig:SFURes}
    \end{subfigure}
    \begin{subfigure}[b]{\linewidth}
        \centering
        \caption{8x50 F1 = 0.960}
        \includegraphics[width=.9\linewidth]{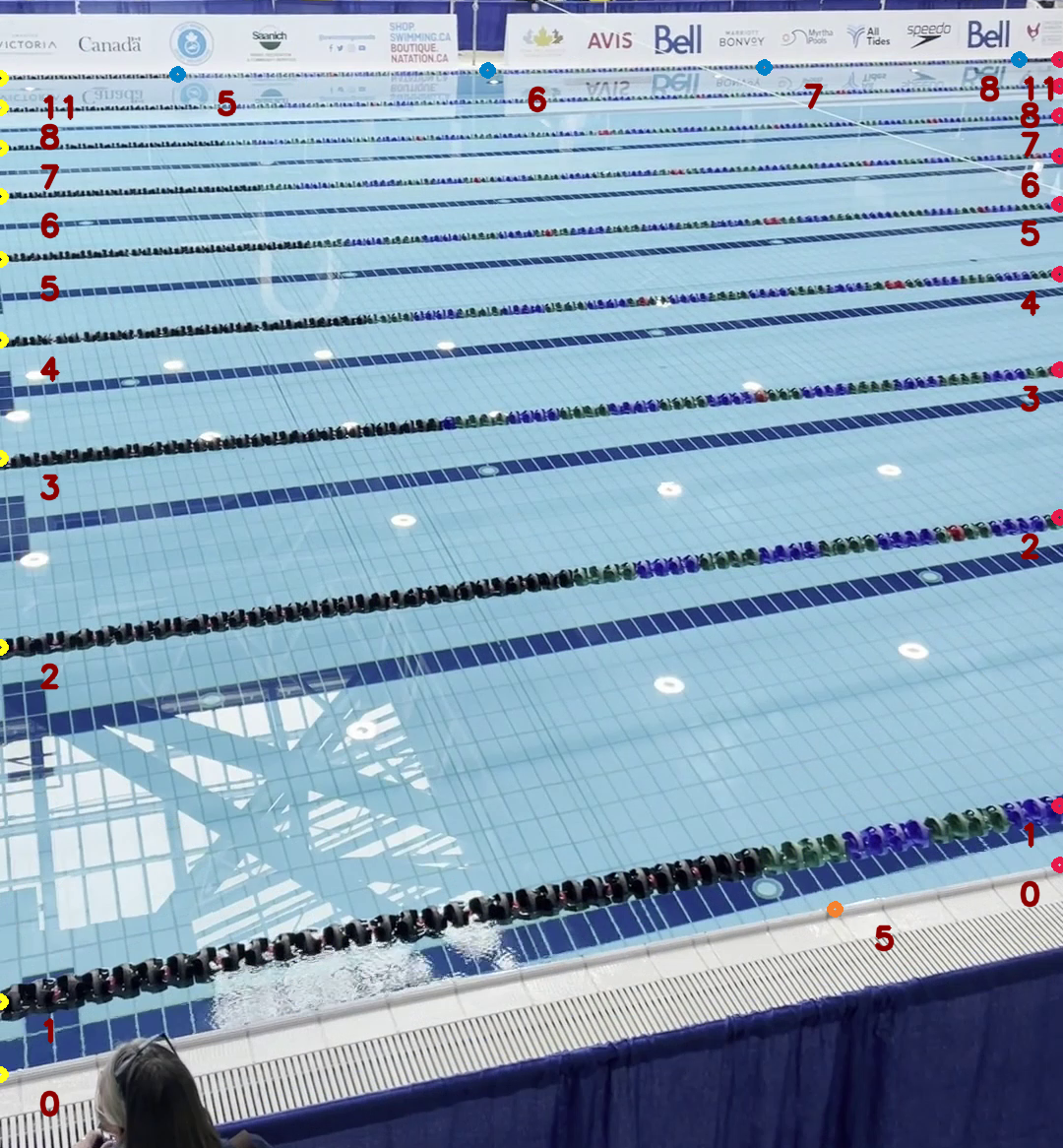}
        \label{fig:SaanichRes}
    \end{subfigure}
    \begin{subfigure}[b]{\linewidth}
        \centering
        \includegraphics[width=\linewidth]{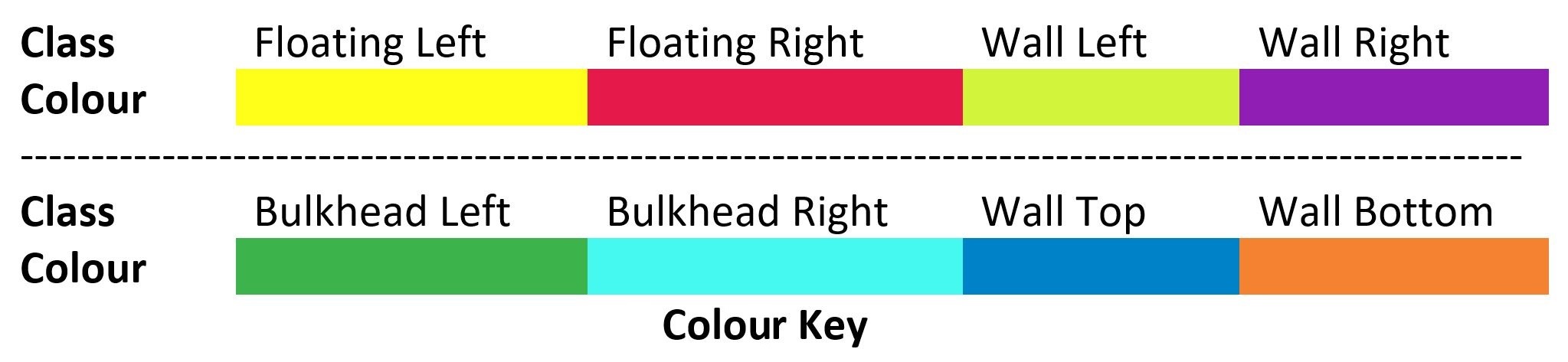}
        \label{fig:keyPointColourKey}
    \end{subfigure}
\end{figure}

\begin{figure}[ht] \ContinuedFloat
    \centering
    \begin{subfigure}[b]{\linewidth} 
        \centering
        \caption{20x25 Left F1 = 0.529}
        \includegraphics[width=\linewidth]{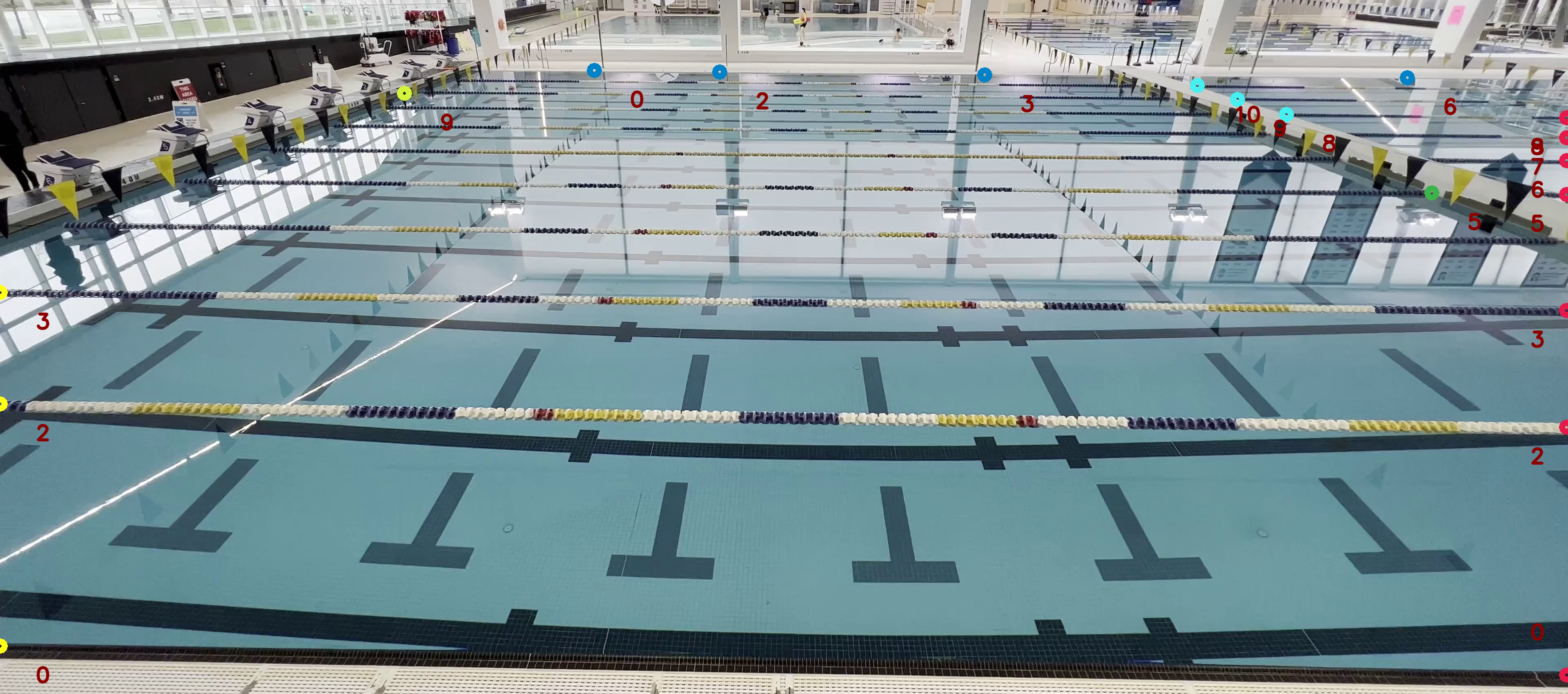}
        \label{fig:UBCLeftRes}
    \end{subfigure}
    \begin{subfigure}[b]{\linewidth}
        \centering
        \caption{20x25 F1 = 0.477}
        \includegraphics[width=\linewidth]{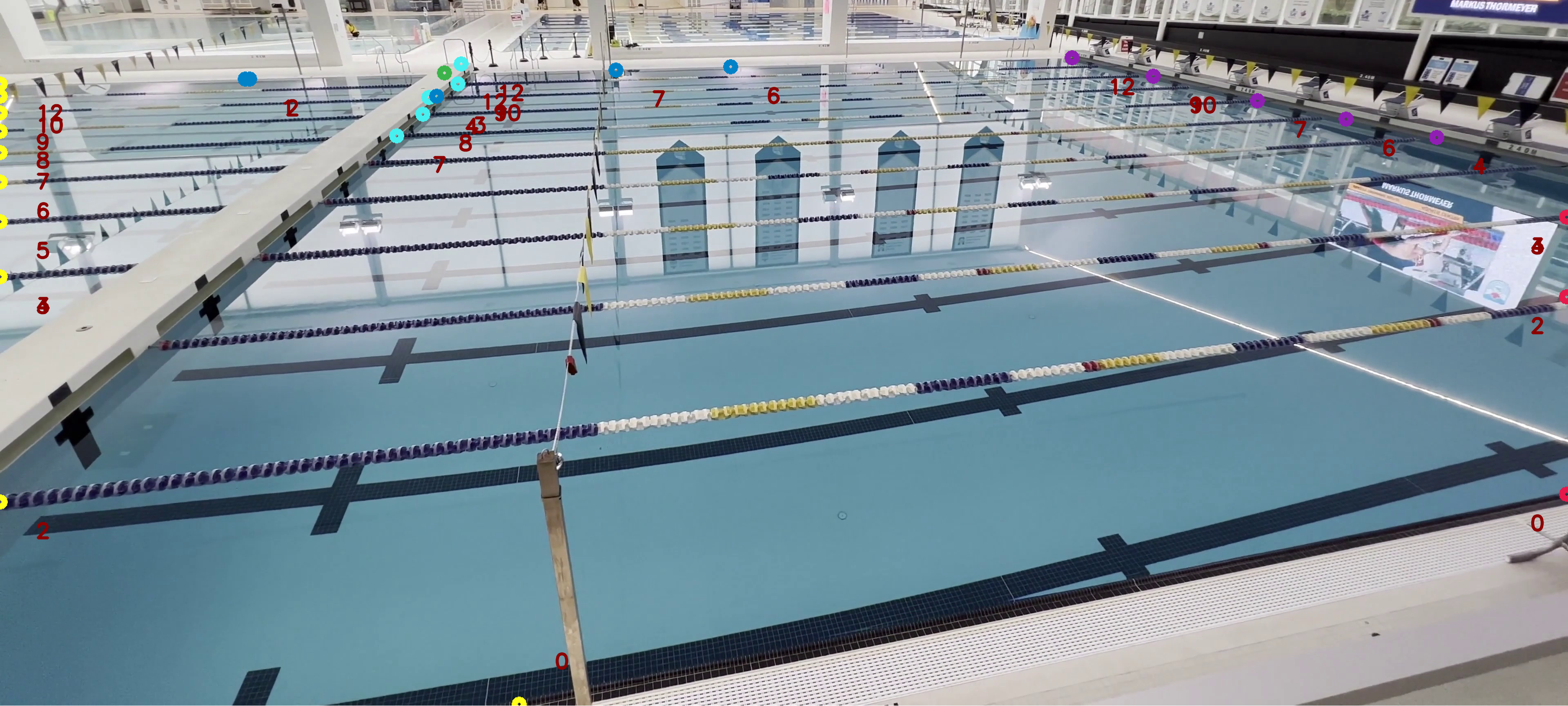}
        \label{fig:UBCRightRes}
    \end{subfigure}
    \caption{Visual results of the key-point detector}
    \label{fig:visualResults}
\end{figure}

\subsection{Discussion} \label{subsec:discussion}
In this section, first, we discuss the results of training the key-point detector. Then, we look at the estimated detectability of the proposed pool key-point model by observing how well the proposed detector trains and performs on the collected pool sequences.

Observing the accuracy as a function of epochs in Figure~\ref{fig:accuracyPlot}, there is no sign of a performance decrease due to over-fitting. While this may be true, the overall accuracy is reasonably low and the test accuracy is much higher than the training accuracy. This is uncommon, as generally, the training data does better than the testing data. The reason for this large difference in values is likely because some pools are easier to detect key-points in than others. In particular, it seems that six-lane pools are easier to detect key-points from than other pools because they have fewer key-points to detect; in addition, the pool can fit into more of the field of view. In comparison, a 16x25 pool which has more key-points and a much smaller fraction of the pool fits in the field of view. Referring to Table~\ref{tab:dataDistabution}, roughly 20\% of the testing sequences are from a six-lane pool, while 8\% of the training is from a six-lane pool. The same phenomenon is observed with 8x50 pools, which have no bulkheads, compared to 16x25 pools which have bulkheads, and thus more key-points to detect.


Figure~\ref{fig:betaAccuracyPlot} displays the F1 Score of the model as a function of $\beta$, the control parameter selecting how confident the detector must be for a key-point prediction to be considered predicted. What this plot shows is that at roughly $0.1 < \beta < 0.995$ the model predicts mostly the same, quantitatively speaking. This means that any values of $\beta > 0.995$ or $\beta < 0.1$ would result in less optimal results, in terms of an F1 score. Qualitatively speaking, while the F1 performance of the model seems to be roughly the same for $0.1 < \beta < 0.995$, when $\beta$ is higher there tends to be larger precision values in exchange for lower recall values, and when $\beta$ is lower the opposite is observed. This result is to be expected as a higher $\beta$ corresponds to a higher entropy which means less certainty about the given prediction.

Figure~\ref{fig:accuracyPlot} shows that after a pixel tolerance of five pixels the F1 score changes at a roughly different rate. This may indicate that the five pixel tolerance suggested in~\cite{citraro2020real} is indeed a good value to choose as the tolerance for measuring key-point prediction quality. 

In Table~\ref{tab:keyPointResults}, the detector performs the best on most floating key-points, does reasonably well with wall left/right/bottom key-points, and struggles with the rest. There may be a few reasons for this.

Firstly, the lack of performance for wall bottom key-points is understandable. The wall closest to the camera gets the smallest fractional field of view. This is due to the physics of cameras in general, that being, the farther something is away from the camera, the more of that thing that can be captured by such camera. Accordingly, there is very little context for the model to reason about the location and class of the key-points present. Annotation of such key-points is easier as the annotator can use temporal knowledge to find key-points. Furthermore, this may indicate that detectability is related to the relative fraction of the pool observed in the image. As such the detectability of wall bottom key-points is low. 

Higher performance was expected for the bulkhead points. Observe Figure~\ref{fig:UBCRightRes}, for which the bulkhead right key-points are detected less accurately. These key-points should be easy to detect. Observing the training data from a similar viewpoint, the bulkhead line that makes the intersection with the lane-ropes is almost always closely accompanied by the flags that cross the pool, this likely impeded the detector. To show that the bulkhead can be detected, Figure~\ref{fig:UBCLeftRes} gives examples of the model detecting bulkhead points. It also seems the model may have had trouble with the bulkheads due to a lack of data. Given adequate data, the detectability is higher than the wall bottom key-points.

Overall wall top key-points were detected poorly. Having an F1 score of $0$, $0.4357$, and $0.2200$ for six, eight, and ten lanes respectively, is not good performance. However, in one sequence, they were detected well. Observe Figure~\ref{fig:SaanichRes}, in which the wall top key-points are detected well. The practiced eye will note that the lane-ropes give markers that sometimes guide the location of the wall top and bottom key-points. The pool in Figure~\ref{fig:SaanichRes} has lane-rope markers that were very well placed. As such it seems the model was able to pick up on these placements. Another explanation for the good wall top key-point detection is that the training data from this pool had double samples from the same location. As such, the model was trained on the same viewpoint as the test sequence. This may indicate that wall top key-points are more affected by changes in viewpoints compared to floating and wall left/right key-points.

Wall right and left key-points should be reasonably easy to detect, and for the most part, they are. The model can confuse them with bulkhead points, however, that is not observed often. The main lack of performance in the wall left and right key-points is due to ordering mismatches, occlusions, and choosing the same point as different key-points. An example can be seen in Figure~\ref{fig:UBCRightRes} for which the true lane five wall right key-point is marked as key-point six and as such the rest of the key-points are incorrect.

Floating left and right points are by far the most reliably detected. Intuitively they should be the easiest to detect as their points are the result of the most dominant lines in the image intersecting with the edge of the frame. As was mentioned in the comment about the wall left and right points, their ordering can sometimes get mismatched and as a result, an entire frame can be detected incorrectly. However, overall they are very detectible key-points. 

The ``bumper key-points'', which are defined as the key-points resulting from the lane-ropes separating the outside lanes from the pool walls, were accounted for reasonably well. When a pool had bumpers the model was able to detect their existence and account for them in the key-point ordering for the floating and wall left/right key-points. This is very important for the creation of a pool homography because the existence of bumpers or lack thereof, can change the perceived length of the pool. An example of the model noticing bumpers can be seen in Figure~\ref{fig:SFURes}, and~\ref{fig:SaanichRes}.

Overall in Figure~\ref{fig:visualResults} the model seems to mainly lose performance due to ordering key-points. That is if the model misses a lane's key-point, it is more likely the rest of the lanes are incorrect. The other cause of miss detection is simply placing key-points in the wrong place or not finding them at all. This may suggest the model architecture of the key-point detector may need to be changed allowing for a model to properly reason about the relative locations of the key-points. It might also suggest that it needs more of the pool to properly reason about what is going on.




\section{Conclusion} \label{sec:conclusion}
The purpose of this work is to create a starting point for the registration of partial pool images. To achieve this task the detection of key-points was proposed for which a homography can be built, as seen in Figure~\ref{fig:exampleHomography}. A pool model was proposed which defines key-points within a general pool. After training a basic key-point detector to detect the proposed key-points, all but the wall bottom key-points, which are still recommended, seem like reasonable locations for key-points. Beyond the proposed key-points in this work, there do not seem to be many more well-defined locations for key-points in an image of a pool. For further work on this subject, a better detector should be considered. Model architectures that increase the receptive field of the model must be tested. More data augmentation methods should be employed. The target function should be made more sophisticated such that the learning algorithm is rewarded for putting key-points close to the ground truth. Higher, image resolutions should be used in training. Lastly, other methods of field localization, considered in section~\ref{sec:relatedWork} should be explored and compared in a meaningful manner to the results found in this work.

\begin{acks}
Thank you to Own the Podium for coordinating like minds in this research area and for funding the Automated Swimming Analytics project of which this research was a part. Thanks to Swimming Canada for supporting this research and for helping organize the collection of pool footage which made this work possible.
\end{acks}

\bibliographystyle{ACM-Reference-Format}
\balance
\bibliography{tw_bibFile}

\appendix

\onecolumn
\section{Detailed Key-Point Results} \label{sec:AllData}


\begin{longtable}{|p{.14\linewidth}|p{.05\linewidth}|p{.05\linewidth}|p{.08\linewidth}|p{.05\linewidth}|p{.05\linewidth}|p{.08\linewidth}|p{.05\linewidth}|p{.05\linewidth}|p{.08\linewidth}|} 
    
	\hline
    \multicolumn{10}{|c|}{\textbf{Table Of key-point Detection Accuracy for Each key-point}}\\
    \hline
    \multicolumn{1} {|r|} {\textbf{Pool Type}} & \multicolumn{3} {|c|} {\textbf{6x25 $\beta=0.15$}} & \multicolumn{3} {|c|} {\textbf{16x25 and 8x50 $\beta=0.90$}} & \multicolumn{3} {|c|} {\textbf{20x25 $\beta=0.70$}}\\
    \hline
    \textbf{key-point} & \textbf{Prec} & \textbf{Rec} & \textbf{Total KPs} & \textbf{Prec} & \textbf{Rec} & \textbf{Total KPs} & \textbf{Prec} & \textbf{Rec} & \textbf{Total KPs}\\
    \hline
    \endfirsthead
    
    \hline
    \multicolumn{10}{|c|}{Continuation of Table Of key-point Detection Accuracy for Each key-point}\\
    \hline
    \multicolumn{1} {|r|} {\textbf{Pool Type}} & \multicolumn{3} {|c|} {\textbf{6x25 $\beta=0.15$}} & \multicolumn{3} {|c|} {\textbf{16x25 and 8x50 $\beta=0.90$}} & \multicolumn{3} {|c|} {\textbf{20x25 $\beta=0.70$}}\\
    \hline
    \textbf{key-point} & \textbf{Prec} & \textbf{Rec} & \textbf{Total KPs} & \textbf{Prec} & \textbf{Rec} & \textbf{Total KPs} & \textbf{Prec} & \textbf{Rec} & \textbf{Total KPs}\\
    \hline
    \endhead
    
    \hline
    \endfoot
    
    \hline
    \multicolumn{10}{| c |}{End of Table}\\
    \hline
    \caption{This table provides the precision and recall of all the key-point detections for all 96 key-points that can be predicted by the model, as well as the number of possible detections in the ground truth. The precision and recall were calculated by assuming that predicted key-points were correct if they were within 5 pixels of the ground truth.\label{tab:allData}}\\
    \endlastfoot
    
    
    Wall Left 0 & 0 & 0 & 3 & 0.2000 & 1.0000 & 10 & 0 & 0 & 2 \\
    \hline
    Wall Left 1 & 0 & 0 & 3 & 0.2727 & 1.0000 & 11 & - & - & 0 \\
    \hline
    Wall Left 2 & 1.0000 & 0.5000 & 6 & 0.8947 & 0.9444 & 19 & 0 & 0 & 6 \\
    \hline
    Wall Left 3 & 0 & 0 & 9 & 1.0000 & 0.9615 & 25 & 0 & 0 & 9 \\
    \hline
    Wall Left 4 & 0 & 0 & 11 & 0.9032 & 0.9032 & 31 & 0 & 0 & 13 \\
    \hline
    Wall Left 5 & 0.4000 & 0.1538 & 13 & 0.8000 & 0.7000 & 35 & 0.0625 & 0.1667 & 16 \\
    \hline
    Wall Left 6 & 0 & 0 & 16 & 0.3590 & 0.4667 & 39 & 0 & 0 & 17 \\
    \hline
    Wall Left 7 & - & - & 0 & 0.5750 & 0.8519 & 40 & 0 & 0 & 18 \\
    \hline
    Wall Left 8 & - & - & 0 & 0.7000 & 0.8750 & 40 & 0.1905 & 0.2353 & 21 \\
    \hline
    Wall Left 9 & - & - & 0 & - & - & 0 & 0.4800 & 0.9231 & 25 \\
    \hline
    Wall Left 10 & - & - & 0 & - & - & 0 & 0.1852 & 0.6250 & 27 \\
    \hline
    Wall Left 11 & 0 & 0 & 18 & 0.4103 & 1.0000 & 39 & - & - & 0 \\
    \hline
    Wall Left 12 & 0 & 0 & 18 & 0.1364 & 0.4286 & 22 & 0.1034 & 1.0000 & 29 \\
    \hline
    Wall Right 0 & 0 & 0 & 4 & 0.7000 & 1.0000 & 10 & 0 & 0 & 5 \\
    \hline
    Wall Right 1 & 0 & 0 & 5 & 0.1538 & 1.0000 & 13 & - & - & 0 \\
    \hline
    Wall Right 2 & 0 & 0 & 9 & 1.0000 & 0.8400 & 21 & 0.2500 & 1.0000 & 8 \\
    \hline
    Wall Right 3 & 0 & 0 & 11 & 1.0000 & 0.9355 & 29 & 0.2727 & 0.3333 & 11 \\
    \hline
    Wall Right 4 & 0.5000 & 0.0769 & 13 & 1.0000 & 0.2500 & 11 & 0.5000 & 0.4375 & 14 \\
    \hline
    Wall Right 5 & 0 & 0 & 15 & 0.4048 & 0.3696 & 42 & 0.7778 & 0.7778 & 18 \\
    \hline
    Wall Right 6 & 1.0000 & 0.2500 & 16 & 0.8511 & 0.8000 & 47 & 0.2632 & 0.2500 & 19 \\
    \hline
    Wall Right 7 & - & - & 0 & 0.6939 & 0.6182 & 49 & 0.7826 & 0.6667 & 23 \\
    \hline
    Wall Right 8 & - & - & 0 & 0.8600 & 0.9348 & 50 & 0.2400 & 0.2727 & 25 \\
    \hline
    Wall Right 9 & - & - & 0 & - & - & 0 & 0.3333 & 0.3226 & 30 \\
    \hline
    Wall Right 10 & - & - & 0 & - & - & 0 & 0.5938 & 0.5938 & 32 \\
    \hline
    Wall Right 11 & 0 & 0 & 17 & 0.3182 & 0.9333 & 44 & - & - & 0 \\
    \hline
    Wall Right 12 & 1.0000 & 0.1176 & 17 & 0.8333 & 1.0000 & 30 & 0.8750 & 1.0000 & 32 \\
    \hline
    Floating Left 0 & 0.9020 & 0.8679 & 53 & 0.9012 & 0.8639 & 162 & 0.8500 & 0.7727 & 40 \\
    \hline
    Floating Left 1 & 1.0000 & 0.5660 & 53 & 0.3168 & 0.9623 & 161 & - & - & 0 \\
    \hline
    Floating Left 2 & 0.9400 & 0.9400 & 50 & 0.8052 & 0.7799 & 154 & 0.9750 & 0.9070 & 40 \\
    \hline
    Floating Left 3 & 0.9583 & 0.9787 & 47 & 0.8667 & 0.8609 & 150 & 0.9167 & 0.8250 & 36 \\
    \hline
    Floating Left 4 & 0.8667 & 0.8667 & 45 & 0.8000 & 0.8056 & 145 & 0.7879 & 0.8387 & 33 \\
    \hline
    Floating Left 5 & 0.8500 & 0.7907 & 43 & 0.8652 & 0.8592 & 141 & 0.8966 & 0.9286 & 29 \\
    \hline
    Floating Left 6 & 0.7561 & 0.7750 & 40 & 0.9051 & 0.9118 & 137 & 0.8276 & 0.9231 & 29 \\
    \hline
    Floating Left 7 & - & - & 0 & 0.9191 & 0.9398 & 136 & 0.7500 & 0.7500 & 28 \\
    \hline
    Floating Left 8 & - & - & 0 & 0.9213 & 0.9590 & 127 & 0.7727 & 0.6800 & 22 \\
    \hline
    Floating Left 9 & - & - & 0 & - & - & 0 & 0.6500 & 0.9286 & 20 \\
    \hline
    Floating Left 10 & - & - & 0 & - & - & 0 & 0.8421 & 0.8421 & 19 \\
    \hline
    Floating Left 11 & 1.0000 & 0.9737 & 38 & 0.2931 & 0.9189 & 116 & - & - & 0 \\
    \hline
    Floating Left 12 & 1.0000 & 1.0000 & 38 & 0.9383 & 0.9744 & 81 & 0.7895 & 0.8824 & 19 \\
    \hline
    Floating Right 0 & 0.9800 & 0.9423 & 52 & 0.8834 & 0.9057 & 163 & 0.7561 & 0.8378 & 41 \\
    \hline
    Floating Right 1 & 1.0000 & 0.5294 & 51 & 0.4313 & 0.9718 & 160 & - & - & 0 \\
    \hline
    Floating Right 2 & 0.9574 & 0.9574 & 47 & 0.7516 & 0.7372 & 153 & 0.8500 & 0.9189 & 40 \\
    \hline
    Floating Right 3 & 1.0000 & 1.0000 & 45 & 0.8844 & 0.8966 & 147 & 0.8378 & 0.9118 & 37 \\
    \hline
    Floating Right 4 & 0.9767 & 0.9767 & 43 & 0.8865 & 0.8929 & 141 & 0.6774 & 0.8077 & 31 \\
    \hline
    Floating Right 5 & 0.9512 & 0.9512 & 41 & 0.9185 & 0.9254 & 135 & 0.7586 & 0.9167 & 29 \\
    \hline
    Floating Right 6 & 0.8750 & 0.8750 & 40 & 0.8855 & 0.8923 & 131 & 0.7500 & 0.8077 & 28 \\
    \hline
    Floating Right 7 & - & - & 0 & 0.8346 & 0.9217 & 127 & 0.8750 & 1.0000 & 24 \\
    \hline
    Floating Right 8 & - & - & 0 & 0.9407 & 0.9823 & 118 & 0.5714 & 0.7500 & 21 \\
    \hline
    Floating Right 9 & - & - & 0 & - & - & 0 & 0.8125 & 0.8667 & 16 \\
    \hline
    Floating Right 10 & - & - & 0 & - & - & 0 & 0.6250 & 0.9091 & 16 \\
    \hline
    Floating Right 11 & 1.0000 & 0.9744 & 39 & 0.3153 & 1.0000 & 111 & - & - & 0 \\
    \hline
    Floating Right 12 & 1.0000 & 1.0000 & 39 & 0.9589 & 0.8974 & 73 & 0.5714 & 0.8889 & 14 \\
    \hline
    Bulkhead Left 0 & - & - & 0 & 0 & 0 & 37 & 0 & 0 & 3 \\
    \hline
    Bulkhead Left 1 & - & - & 0 & 0 & 0 & 37 & 0 & 0 & 3 \\
    \hline
    Bulkhead Left 2 & - & - & 0 & 0.1944 & 1.0000 & 36 & 0 & 0 & 15 \\
    \hline
    Bulkhead Left 3 & - & - & 0 & 0.0294 & 0.0312 & 34 & 0.2000 & 0.5714 & 20 \\
    \hline
    Bulkhead Left 4 & - & - & 0 & 0 & 0 & 34 & 0.0357 & 0.1000 & 28 \\
    \hline
    Bulkhead Left 5 & - & - & 0 & 0.3235 & 0.4074 & 34 & 0.5000 & 1.0000 & 30 \\
    \hline
    Bulkhead Left 6 & - & - & 0 & 0.0303 & 0.5000 & 33 & 0 & 0 & 34 \\
    \hline
    Bulkhead Left 7 & - & - & 0 & 0.0303 & 0.1429 & 33 & 0.3514 & 1.0000 & 37 \\
    \hline
    Bulkhead Left 8 & - & - & 0 & 0.0303 & 1.0000 & 33 & 0 & 0 & 39 \\
    \hline
    Bulkhead Left 9 & - & - & 0 & - & - & 0 & 0.2683 & 1.0000 & 41 \\
    \hline
    Bulkhead Left 10 & - & - & 0 & - & - & 0 & 0.1667 & 0.7778 & 42 \\
    \hline
    Bulkhead Left 11 & - & - & 0 & 0 & 0 & 33 & - & - & 0 \\
    \hline
    Bulkhead Left 12 & - & - & 0 & 0 & 0 & 33 & 0 & 0 & 46 \\
    \hline
    Bulkhead Right 0 & - & - & 0 & 0 & 0 & 38 & 0 & 0 & 7 \\
    \hline
    Bulkhead Right 1 & - & - & 0 & 0 & 0 & 38 & 0 & 0 & 7 \\
    \hline
    Bulkhead Right 2 & - & - & 0 & 0 & 0 & 37 & 0 & 0 & 17 \\
    \hline
    Bulkhead Right 3 & - & - & 0 & 0 & 0 & 36 & 0.1304 & 0.1429 & 23 \\
    \hline
    Bulkhead Right 4 & - & - & 0 & 0 & 0 & 35 & 0 & 0 & 27 \\
    \hline
    Bulkhead Right 5 & - & - & 0 & 0 & 0 & 35 & 0 & 0 & 32 \\
    \hline
    Bulkhead Right 6 & - & - & 0 & 0 & 0 & 35 & 0.1429 & 1.0000 & 35 \\
    \hline
    Bulkhead Right 7 & - & - & 0 & 0.2424 & 1.0000 & 33 & 0.4474 & 0.8500 & 38 \\
    \hline
    Bulkhead Right 8 & - & - & 0 & 0 & 0 & 33 & 0.7381 & 0.8857 & 42 \\
    \hline
    Bulkhead Right 9 & - & - & 0 & - & - & 0 & 0.4524 & 0.7600 & 42 \\
    \hline
    Bulkhead Right 10 & - & - & 0 & - & - & 0 & 0.4773 & 0.6176 & 44 \\
    \hline
    Bulkhead Right 11 & - & - & 0 & 0 & 0 & 33 & - & - & 0 \\
    \hline
    Bulkhead Right 12 & - & - & 0 & 0 & 0 & 33 & 0.2045 & 0.4737 & 44 \\
    \hline
    Wall Top 0 & 0 & 0 & 18 & 0.4000 & 0.7368 & 35 & 0.6667 & 0.6875 & 33 \\
    \hline
    Wall Top 1 & 0 & 0 & 16 & 0.3824 & 1.0000 & 34 & 0.1429 & 0.2273 & 35 \\
    \hline
    Wall Top 2 & - & - & 0 & 0.2581 & 0.5714 & 31 & 0 & 0 & 37 \\
    \hline
    Wall Top 3 & - & - & 0 & 0.2558 & 0.6471 & 43 & 0.4146 & 0.5000 & 41 \\
    \hline
    Wall Top 4 & - & - & 0 & 0.2895 & 1.0000 & 38 & - & - & 0 \\
    \hline
    Wall Top 5 & - & - & 0 & 0.2340 & 1.0000 & 47 & 0.0476 & 0.2500 & 42 \\
    \hline
    Wall Top 6 & - & - & 0 & 0.3846 & 0.7500 & 39 & 0.3250 & 0.2955 & 40 \\
    \hline
    Wall Top 7 & 0 & 0 & 21 & 0.2250 & 0.6923 & 40 & 0 & 0 & 39 \\
    \hline
    Wall Top 8 & 0 & 0 & 19 & 0.2791 & 0.7059 & 43 & 0 & 0 & 34 \\
    \hline
    Wall Bottom 0 & 0 & 0 & 5 & 0.2000 & 0.4286 & 15 & 0 & 0 & 13 \\
    \hline
    Wall Bottom 1 & 0 & 0 & 13 & 0 & 0 & 21 & 0 & 0 & 16 \\
    \hline
    Wall Bottom 2 & - & - & 0 & 0.2000 & 0.8571 & 30 & 0 & 0 & 16 \\
    \hline
    Wall Bottom 3 & - & - & 0 & 0 & 0 & 41 & - & - & 0 \\
    \hline
    Wall Bottom 4 & - & - & 0 & 0 & 0 & 12 & - & - & 0 \\
    \hline
    Wall Bottom 5 & - & - & 0 & 0.2549 & 0.2889 & 51 & 0 & 0 & 13 \\
    \hline
    Wall Bottom 6 & - & - & 0 & 0.2250 & 0.2647 & 40 & 0 & 0 & 18 \\
    \hline
    Wall Bottom 7 & 0 & 0 & 20 & 0.2692 & 0.4375 & 26 & 0 & 0 & 10 \\
    \hline
    Wall Bottom 8 & 0 & 0 & 26 & 0 & 0 & 15 & - & - & 0 \\
    \hline
    
\end{longtable}
\twocolumn

\end{document}